\documentclass[10pt,sigconf,natbib=false]{acmart}
\pdfoutput=1
\AtBeginDocument{%
  }

\usepackage{graphicx}
\usepackage{float}
\usepackage{subfig}
\usepackage{multirow}
\usepackage{tabularx}
\usepackage{tikz}
\usepackage{hyperref}
\usepackage{adjustbox}
\acmConference[MMSys'24]{MMSys'24}{April, 15-18, 2024}{Bari, Italy}

\usepackage{etoolbox}
\makeatletter
\patchcmd{\authornote}{\g@addto@macro\addresses{\@authornotemark}}{}{}{}
\makeatother

\RequirePackage[
  datamodel=acmdatamodel,
  style=acmnumeric,sorting=none
  ]{biblatex}

\DeclareUnicodeCharacter{2218}{\circ}

\begin{document}

\title{Panonut360: A  Head and Eye Tracking Dataset \\ for  Panoramic Video}
\renewcommand{\shorttitle}{Panonut360: A  Head and Eye Tracking Dataset for  Panoramic Video}

\author{Yutong Xu$^{\dag}$, Junhao Du$^{\dag}$, Jiahe Wang, Yuwei Ning, Sihan Zhou and Yang Cao$^{\ast}$}
\authornote{Yang Cao is the corresponding author.}
\authornote{Both authors contributed equally to this research.}
\affiliation{
  \institution{School of Electronic Information and Communications, Huazhong University of Science and Technology}
  \city{Wuhan}
  \country{China}
}
\email{{yukon_hsu@, junhao_du@, jiahe_wang@, yuwei_ning@, xia9cai@, ycao@}hust.edu.cn}  
 
\renewcommand{\shortauthors}{Y. Xu et al.}

\begin{abstract}
With the rapid development and widespread application of VR/AR technology, maximizing the quality of immersive panoramic video services that match users’ personal preferences and habits has become a long-standing challenge. Understanding the saliency region where users focus, based on data collected with HMDs (Head-mounted Displays), can promote multimedia encoding, transmission, and quality assessment. At the same time, large-scale datasets are essential for researchers and developers to explore short/long-term user behavior patterns and train AI models related to panoramic videos. However, existing panoramic video datasets often include low-frequency user head or eye movement data through short-term videos only, lacking sufficient data for analyzing users’ Field of View (FoV) and generating video saliency regions. 

Driven by these practical factors, in this paper, we present a head and eye tracking dataset involving 50 users (25 males and 25 females) watching 15 panoramic videos (mostly in 4K). The dataset provides details on the viewport and gaze attention locations of users. Besides, we present some statistics samples extracted from the dataset. For example, the deviation between head and eye movements challenges the widely held assumption that gaze attention decreases from the center of the FoV following a Gaussian distribution. Our analysis reveals a consistent downward offset in gaze fixations relative to the FoV in experimental settings involving multiple users and videos. That’s why we name the dataset \textit{Panonut}, a saliency weighting shaped like a donut. Finally, we also provide a script that generates saliency distributions based on given head or eye coordinates and pre-generated saliency distribution map sets of each  video from the collected eye tracking data.

The dataset and related code are publicly available on our website: \href{https://dianvrlab.github.io/Panonut360/}{\textit{https://dianvrlab.github.io/Panonut360/}}.

\end{abstract}


\acmYear{2024}\copyrightyear{2024}
\acmConference[MMSys '24]{ACM Multimedia Systems Conference 2024}{April 15--18, 2024}{Bari, Italy}
\acmBooktitle{ACM Multimedia Systems Conference 2024 (MMSys '24), April 15--18, 2024, Bari, Italy}
\acmDOI{10.1145/3625468.3652176}
\acmISBN{979-8-4007-0412-3/24/04}

\maketitle

\section{INTRODUCTION}
With the continuous development of immersive panoramic video production and transmission technology, Virtual Reality (VR) and Augmented Reality (AR) have been applied in fields such as  education\cite{alfalah2018perceptions}, medicine\cite{laver2017virtual}, engineering training\cite{hilfert2016low}, entertainment and gaming\cite{mcmahan2012evaluating}, and have been a key enabling technology for the \textit{metaverse}. The popularity of VR/AR Head Mounted Displays (HMD) has led to a sharp increase in demand for the transmission of panoramic streaming videos. However, since panoramic videos need to be streamed at high quality, up to several times bandwidth resources would be consumed compared to traditional videos, which poses a severe bandwidth challenge.

Panoramic video streaming adaptation based on users’ FoV  helps to decrease bandwidth requirements. To achieve this, panoramic videos are spatially divided into independent video tiles, and bitrate allocation for a specific tile is chosen based on the fact that whether the tile is located within the viewport, the portion of spherical surface focused by the user projected to a segment of plane\cite{jocb}. FoV prediction and video saliency detection are both considered effective methods to improve video bitrate allocation performance. Both of these tasks can be modeled as time series prediction and content saliency detection tasks in AI downstream applications, but they require a comprehensive dataset of user head and eye tracking trajectories, as well as corresponding video saliency data. To our knowledge, there are few high-quality datasets that can meet the requirements for training currently. Due to the limited duration of user data, most existing datasets lack sufficient data to train these models effectively. 

Some researchers and developers have already collected head/gaze tracking data and attention saliency. Table \ref{tab:dataset_comparison} presents the basic information of these datasets, where "Num. Users" and "Num. Contents" present the number of participants and videos, "Length" presents the length of videos and "Freq" presents the sampling frequency of the eye tracker. Although sampling data from videos with short durations ( around 60s ) \cite{rai2017dataset} is friendly for reducing viewer fatigue, it brings difficulties in analyzing general behaviors and habits that users exhibit during prolonged engagement with the content. To address this issue, we followed the former approach of Wu \textit{et al.} \cite{wu2017dataset}, providing participants with sufficient rest and adaptation time, editing some videos shorter to ensure that the content remained engaging for the participants throughout the experiment. Most of previous datasets are sampled at 30-60 Hz, but to obtain more valuable insights, higher sampling rates are necessary, so we collect head/gaze tracking data at a frequency of 120 Hz. Only a few datasets provide methods to generate saliency as we do \cite{xu2018gaze,nguyen2019saliency}.

\begin{table*}[htbp]
  \centering
  \caption{Comparison of Different Dataset}
  \label{tab:dataset_comparison}
  \begin{tabular}{|c|cccc|c|c|c|c|}
    \hline
    \multirow{2}{*}{Dataset} & \multicolumn{4}{c|}{Details} & \multirow{2}{*}{Sal} & \multirow{2}{*}{Head} & \multirow{2}{*}{Gaze} & \multirow{2}{*}{Content} \\ \cline{2-5}
    & Num.Users & Num.Contents  & Length/s & Freq/Hz  & & & &\\ \hline
     Wang \textit{et al.} \cite{wang2022salientvr} & 20 & 22 & 73-1335 &30 & - & $\checkmark$ & $\checkmark$ & Videos  \\
     Jin \textit{et al.} \cite{jin2022you}& 100 & 27 & 60 & 100 & - & $\checkmark$ & $\checkmark$ & Videos  \\
     Wu \textit{et al.} \cite{wu2017dataset}& 48 & 18 & 164-655 & 100 & - & $\checkmark$ & - & Videos \\
     Corbillon \textit{et al.} \cite{corbillon2017360}& 59 & 7 & 70 &30 & $\checkmark$ & $\checkmark$ & - & Videos  \\
     Xu \textit{et al.} \cite{xu2018gaze}& 25 & 208 & 20-70 & 30 & $\checkmark$ & $\checkmark$ & $\checkmark$  & Videos \\
    Rai \textit{et al.} \cite{rai2017dataset}& 40 & 60 & 25 &60 & - & $\checkmark$ & $\checkmark$ & Images  \\
     Nguyen \textit{et al.} \cite{nguyen2019saliency}& 157 & 24 & 60-655 &30-100 &  $\checkmark$ & $\checkmark$ & - & Videos  \\ \hline 
     \multicolumn{9}{c}{} \\ \cline{1-9}
      \textbf{Our dataset} & 50 & 15 & 140-352 & 120 &  $\checkmark$ & $\checkmark$ & $\checkmark$ & Videos  \\ \hline
  \end{tabular}
\end{table*}

To this end, we first conduct measurements on participants' viewing behavior in panoramic scenarios. Afterwards, the collected data are statistically analyzed in order to obtain an accurate distribution of eye and head movements, instead of simply assuming a Gaussian distribution around the head position. The results indicate that the distribution of gaze during long-term video viewing does not follow a Gaussian distribution, instead, it exhibits an anisotropic weighting which is partially shaped like a "donut", which we name it \textit{Panonut}. Finally, we express our appreciation to Nguyen \textit{et al.} for their work, as they have made their data and code for generating saliency publicly accessible. We refer to their work in our own research and provide a script for generating saliency maps from our dataset.

The following paper is organized as follows. Section 2 presents the data collection procedure and describes the dataset. Section 3 presents some preliminary analyses of our dataset. Section 4 gives a simple visualization of saliency data. Section 5 presents the usage of the saliency generation script and visualization. Finally, the paper is concluded in section 6.

\section{DATASET}
\subsection{Data Collection Procedure}
We use the HTC VIVE Pro Eye for data collection and take advantage of the SRanipal SDK\footnote{https://developer-express.vive.com/resources/vive-sense/eye-and-facial-tracking-sdk/} provided by VIVE for tracking both head and eye movements in real time. This device supports sampling data at 120Hz with an error range of 0.5$^{\circ}$-1.1$^{\circ}$, and is capable of tracking eye movements within a range of 110°. We developed our project on Unity based on SteamVR, that allows us to measure head and eye movement data, thereby enabling us to complete our data collection.

Prior to the start of data collection, we conduct interviews to determine if participants have had previous exposure to VR and to ensure their adaptation to the VR scenario. Moreover, we perform a view point calibration to ensure precise data collection. We adjust the tightness of the HMD and the volume of the headphones for each volunteer, allowing them to enjoy an immersive viewing experience of panoramic videos with audio, devoid of any external disturbances. As far as we know, many studies typically use videos without audio. Our dataset may be advantageous for understanding the effects of audio on user behaviour in immersive scenarios. 

Due to the long duration of the data collection process, we allow the participants to take intermittent breaks after each video to prevent the effects of fatigue. After each break, the view point calibration procedures are performed again.

\subsection{Coordinate Description}
 In Figure 1(a), the world coordinate system is set up with the origin at the center of the sphere, while the orientation of the axes is determined using the left-hand rule. Specifically, the $Y$ axis is oriented toward the top of the sphere, and the $X$, $Y$, and $Z$ axes are oriented towards the right, up, and forward directions, respectively. Assuming the user is located at the center of the sphere , watching a panoramic video projected onto the inside of the sphere. A unit vector $H$ is defined to represent the direction of the user's head (or the center of the FoV), with coordinates $(H_{x}, H_{y}, H_{z})$. Additionally, a unit vector $G$ is defined to represent the direction of their gaze (in other words, the fixation point of gaze), with coordinates $(G_{x}, G_{y}, G_{z})$.
\begin{figure}[htbp]
    \centering
    \subfloat[]{\includegraphics[width=0.55\hsize]{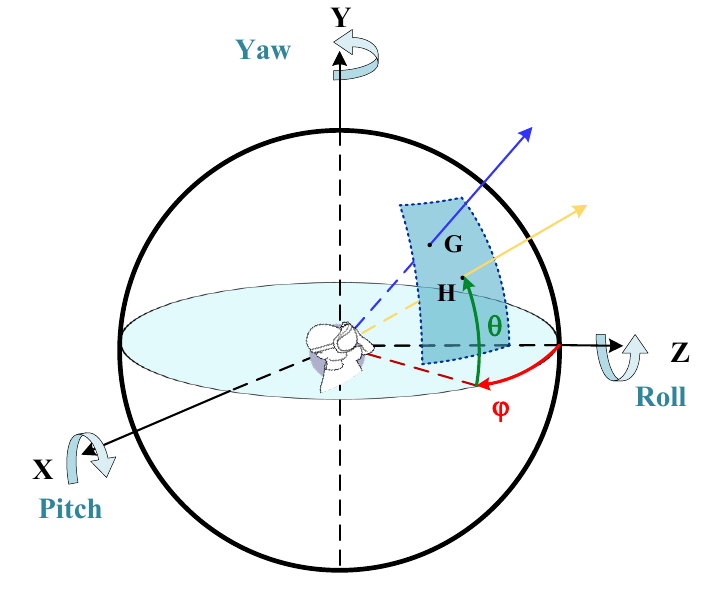}}
    \subfloat[]{\adjustbox{raise=-3.1pt}{\includegraphics[width=0.46\hsize]{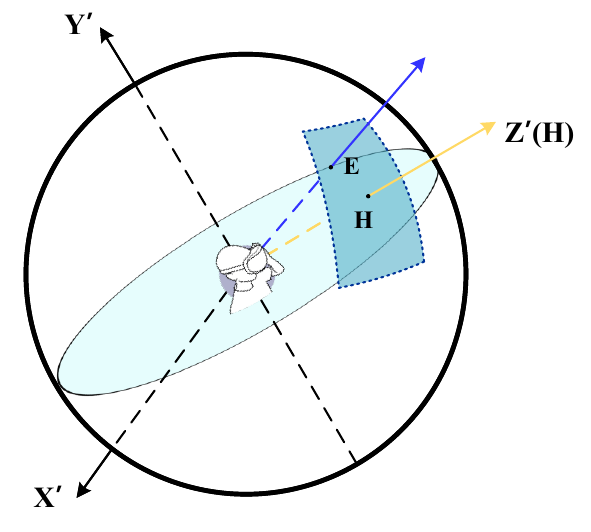}}}
    \caption{(a) Illustrates the world coordinate system, and (b) Illustrates the head coordinate system.}
    \label{coordinate}
\end{figure}

In Figure 1(b), a left-handed coordinate system is set up with  the user's head position as the origin, and the $Z'$ axis pointing in the same direction as  the user's head  orientation (i.e., the unit vector $H$). This coordinate system is called the head coordinate system. In this coordinate system, the coordinate of the vector $E$ is $(E_{x}, E_{y}, E_{z})$, which represents the unit vector $G$ in the head coordinate system. Note that both $E$ and $G$ represent the direction of the user's gaze. $E$ is obtained from the eye tracker on the HMD, while $G$ is obtained from Unity. They are expressed in different coordinate systems, resulting in different numerical values for their coordinates. However, they share the same essence. In addition, $E$ and $G$ are abnormal values when the user blinks.

Longitude $\phi$ is defined as the angle of rotation around the $Y$ axis from the $Z$ axis in a clockwise direction, with a range of $(-\pi$, $\pi]$, while latitude $\theta$ is defined as the angle of deviation from the equator to the $Y$ axis, with a range of $[-\frac{\pi}{2}, \frac{\pi}{2}]$. Using the coordinates of $H$, the following two equations can be used to calculate $\phi$ and $\theta$.

\begin{equation}
    \phi= \left\{
        \begin{aligned}
            \frac{\pi}{2}-arctan(\frac{z}{x}), \quad x \ge 0\\
            -\frac{\pi}{2}-arctan(\frac{z}{x}), \quad x < 0
        \end{aligned}
        \right 
        .
\end{equation}

\begin{equation}
    \theta = arcsin(y)
\end{equation}

\begin{figure}[htbp]
    \centering
    {\includegraphics[width=\hsize]{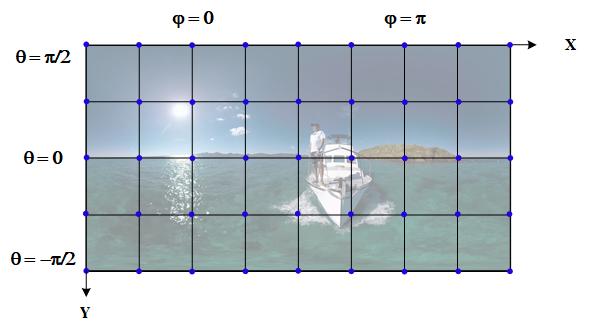}}
    \caption{Schematic diagram of the correspondence between ($\phi$, $\theta$) and ($x$, $y$) coordinates.}
    \label{fig1}
\end{figure}
Figure 2 shows how to project a 3-D video onto a 2-D plane, using the top left corner of the plane serving as the origin of the $X$-$Y$ coordinate system. The pixel coordinates $(x, y)$ are defined, where $x \in [0,Width-1]$ and $y \in [0,Height-1]$. $Width$ and $Height$ represent the width and height of an image, respectively.    The projection equation is as follows:

\begin{equation}
    x= \left\{
        \begin{aligned}
        (\frac{1}{4}+\frac{\phi}{2\pi} ) *Width, \quad \phi \ge -\frac{\pi}{2}\\
        (\frac{5}{4}+\frac{\phi}{2\pi} ) *Width, \quad \phi < -\frac{\pi}{2}
        \end{aligned}
        \right
        .
\end{equation}
\begin{equation}
    y = (\frac{1}{2} - \frac{\theta}{\pi}) * Height
\end{equation}

As is well known, a rotation can be uniquely determined by a quaternion\footnote{https://en.wikipedia.org/wiki/Quaternions\_and\_spatial\_rotation}. Similarly, we provide quaternions to represent $G$ (with components $G_{qw}, G_{qx}, G_{qy}$, and $G_{qz}$) and H (with components $H_{qw}, H_{qx}, H_{qy}$, and $H_{qz}$), whose correspondence with world coordinates can be determined from equation (5).
\begin{equation}
    \left[
\begin{array}{cc}
     x \\
     y \\
     z
\end{array}
\right] = \left[
\begin{array}{cc}
     2*qx*qz+2*qy*qw  \\
     2*qy*qz-2*qx*qw  \\
     1-2*qx^{2}-2*qy^{2}
\end{array}
\right]
\end{equation}

Assuming that the user does not rotate, his forward vector is defined as (0, 0, 1), represented by a unit quaternion of (1, 0, 0, 0). If the user performs a 90° upward rotation (i.e. looks up), the forward vector changes to (0, 1, 0), and the corresponding unit quaternion can be expressed as ($\frac{1}{\sqrt{2}}$, -$\frac{1}{\sqrt{2}}$, 0, 0).

\subsection{Dataset Format}
Our dataset is openly available on our website. The coordinate system of all metrics has been defined in Figure \ref{coordinate}, for each participant and each video, the following metrics are included in our production dataset in CSV format:
\begin{itemize}
    \item PlayerTime (T): the video playback time in (UTC/GMT) in “HH:mm:ss.fff” format.   
    \item Frame : the collected data corresponds to the video frames.
    \item HeadPosition (x, y, z): the coordinate of the HMD device in the Unity world coordinate system correspond to vector $H$.
    \item EyeCombinePosition (x, y, z): the coordinate of of the user's gaze in head coordinate system, which is correspond to vector $E$.
    \item EyePosition (x, y, z): the coordinate of the user's gaze in the Unity world coordinate system, which is correspond to vector $G$.
    \item HeadEular (yaw, pitch, roll): the eular angle of the HMD device in the Unity world coordinate system. If it is necessary to use these parameters, we recommend  to recalculate based on the coordinate system shown in the Figure \ref{coordinate}.
    \item HeadQuaternion (x, y, z, w): the unit quaternion of the HMD device.  
    \item EyeQuaternion (x, y, z, w): the unit quaternion of the user's gaze. 
    
\end{itemize}

Note that \textit{EyeCombinePosition} and \textit{EyePosition} might be abnormal due to participants partially blinking (which is an unavoidable physiological activity) during data collection (in the HTC eye tracker, the gaze coordinates are set to (1, -1, -1) by default during blinking). These abnormal data need to be filtered out.

\subsection{Video Information}

In Table \ref{tab:video info}, we present information about the name, length and content category of the videos selected in the dataset. These videos cover different genres such as natural landscapes, cartoons, sports, stories, and films. Sample images from each video are shown in Figure 3. 
\begin{table}[htbp]
    \centering
        \caption{Videos Information}
      \label{tab:video info}
      \begin{tabular}{llll}
        \toprule
        No. & Video Name & Length & Category\\
        \hline
        1 & Alien & $4^{\prime}53^{\prime\prime}$  & Film\\
        2 & Cooking & $3^{\prime}17^{\prime\prime}$ & Story\\
        3 & CutRope & $2^{\prime}20^{\prime\prime}$ & Cartoon\\
        4 & Dinosaurs & $2^{\prime}39^{\prime\prime}$ & Cartoon\\
        5 & Diving & $6^{\prime}52^{\prime\prime}$ & Sport\\
        6 & Dunkerque & $3^{\prime}19^{\prime\prime}$ & Film\\
        7 & Jordan & $6^{\prime}34^{\prime\prime}$ & Landscape\\
        8 & OculusQuest & $6^{\prime}09^{\prime\prime}$ & Landscape\\
        9 & RollerCoaster & $3^{\prime}26^{\prime\prime}$ & Sport\\
        10 & SpaceWalk & $6^{\prime}23^{\prime\prime}$ & Story\\
        11 & SpongeBob & $5^{\prime}15^{\prime\prime}$ & Cartoon\\
        12 & StarNight & $2^{\prime}54^{\prime\prime}$ & Landscape\\
        13 & Surfing & $3^{\prime}25^{\prime\prime}$ & Sport\\
        14 & UFO &  $2^{\prime}39^{\prime\prime}$ & Film\\
        15 & UnderWater & $5^{\prime}20^{\prime\prime}$ & Landscape\\

        \bottomrule
      \end{tabular}
    \end{table}

\begin{figure}[htbp]

\label{tab:videoframe}

    \begin{minipage}[h]{0.32\linewidth}
    		\vspace{1pt}
    		\centerline{\includegraphics[width=\textwidth]{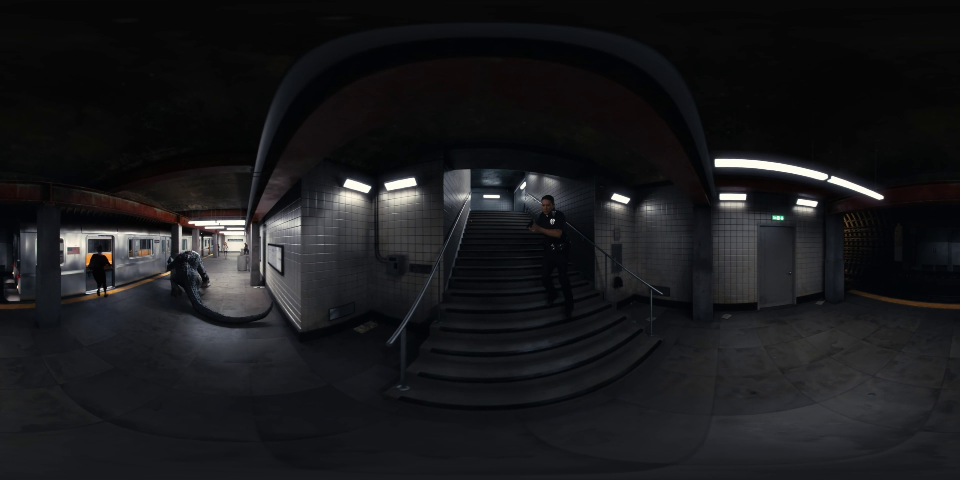}}
    		\centerline{\text{Alien 1min26s}}
    	\end{minipage}
    \begin{minipage}[h]{0.32\linewidth}
    		\vspace{3pt}
    		\centerline{\includegraphics[width=\textwidth]{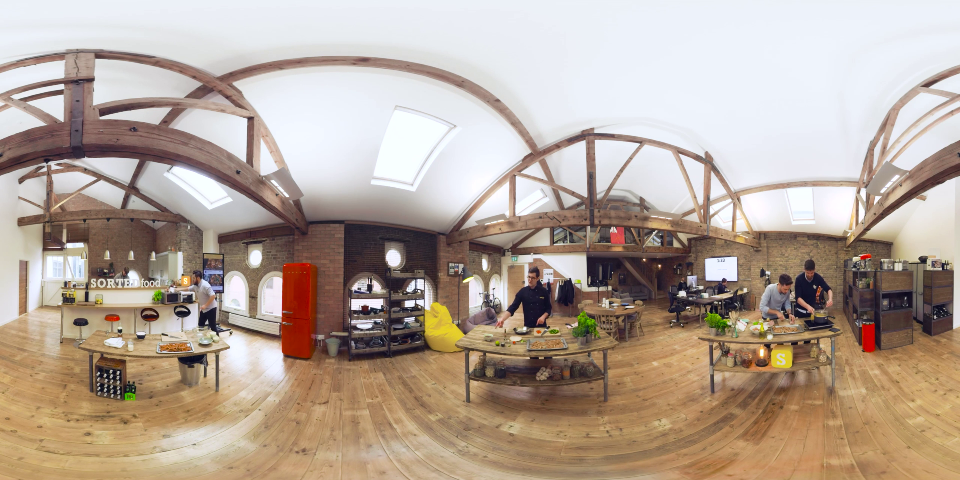}}
    	 
    		\centerline{\text{Cooking 14s}}
    	\end{minipage}
    \begin{minipage}[h]{0.32\linewidth}
    		\vspace{3pt}
    		\centerline{\includegraphics[width=\textwidth]{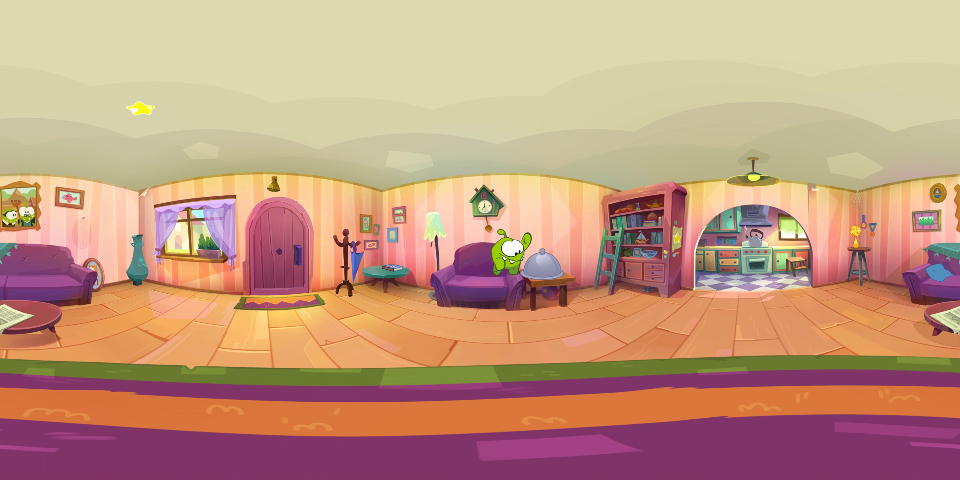}}
    	 
    		\centerline{\text{CutRope 23s}}
    	\end{minipage}
    \begin{minipage}[h]{0.32\linewidth}
    		\vspace{1pt}
    		\centerline{\includegraphics[width=\textwidth]{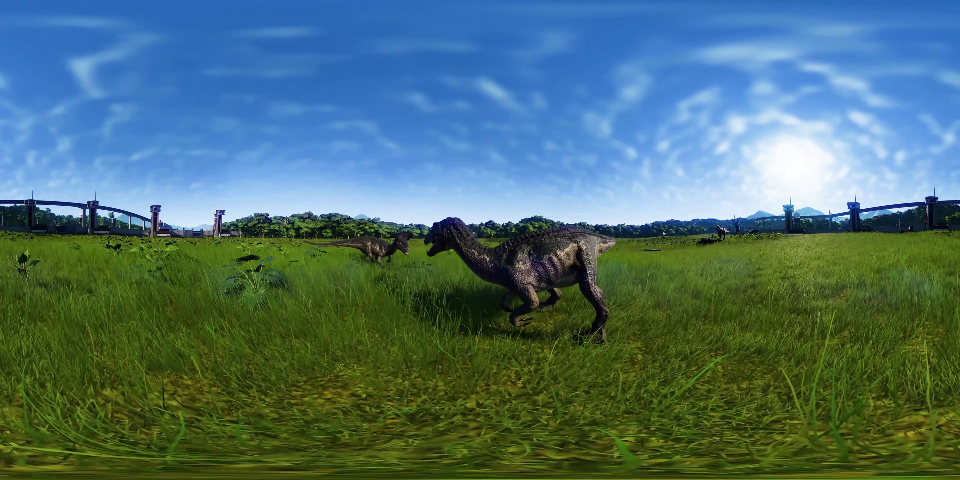}}
    		\centerline{Dinosaurs 22s}
    	\end{minipage}
    \begin{minipage}[h]{0.32\linewidth}
    		\vspace{3pt}
    		\centerline{\includegraphics[width=\textwidth]{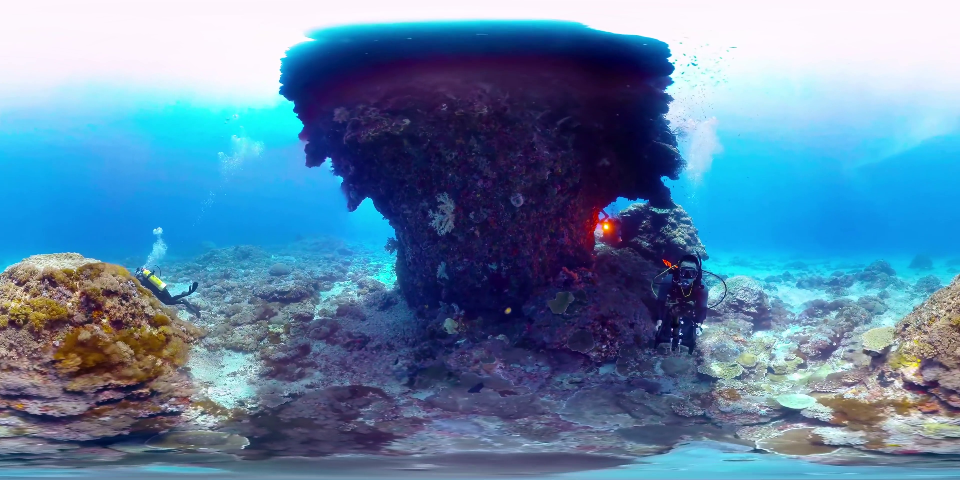}}
    	 
    		\centerline{Diving 2min19s}
    	\end{minipage}
    \begin{minipage}[h]{0.32\linewidth}
    		\vspace{3pt}
    		\centerline{\includegraphics[width=\textwidth]{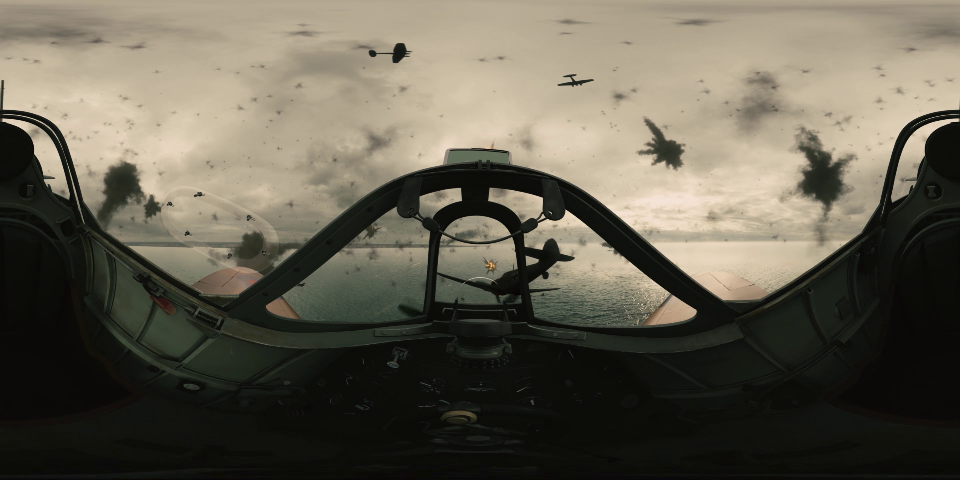}}
    	 
    		\centerline{Dunkerque 3min18s}
    	\end{minipage}
     \begin{minipage}[h]{0.32\linewidth}
    		\vspace{3pt}
    		\centerline{\includegraphics[width=\textwidth]{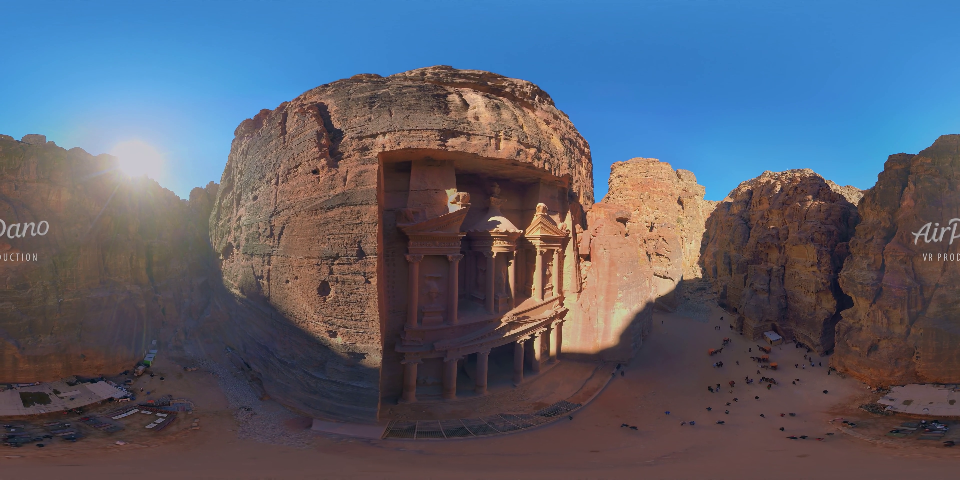}}
    		\centerline{\text{Jordan 3min18}}
    	\end{minipage}
    \begin{minipage}[h]{0.32\linewidth}
    		\vspace{3pt}
    		\centerline{\includegraphics[width=\textwidth]{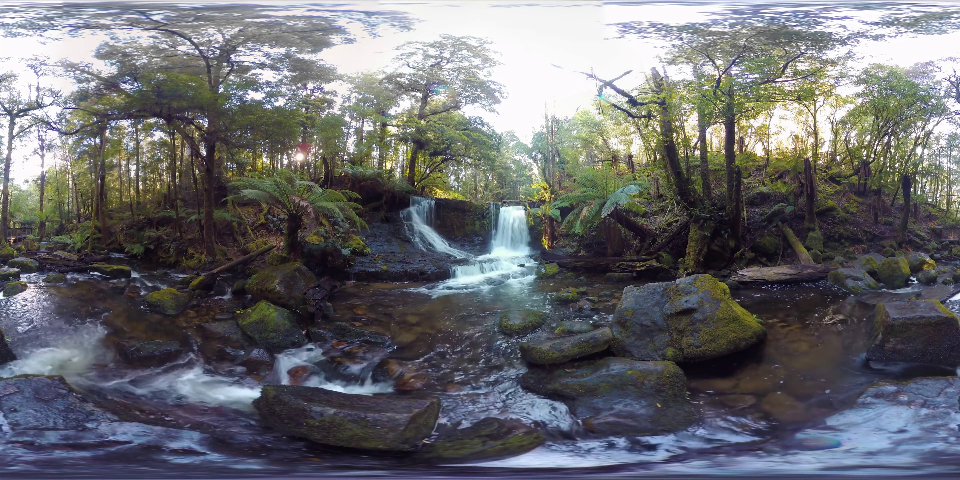}}
    	 
    		\centerline{\text{OculusQuest 22s}}
    	\end{minipage}
    \begin{minipage}[h]{0.32\linewidth}
    		\vspace{1pt}
    		\centerline{\includegraphics[width=\textwidth]{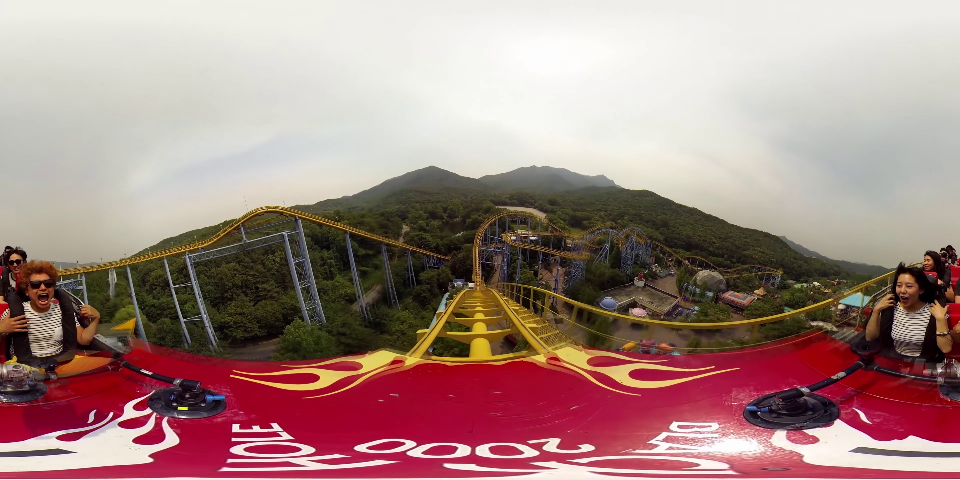}}
    	 
    		\centerline{\text{RollerCoaster 1min53s}}
    	\end{minipage}
     \begin{minipage}[h]{0.32\linewidth}
    		\vspace{3pt}
    		\centerline{\includegraphics[width=\textwidth]{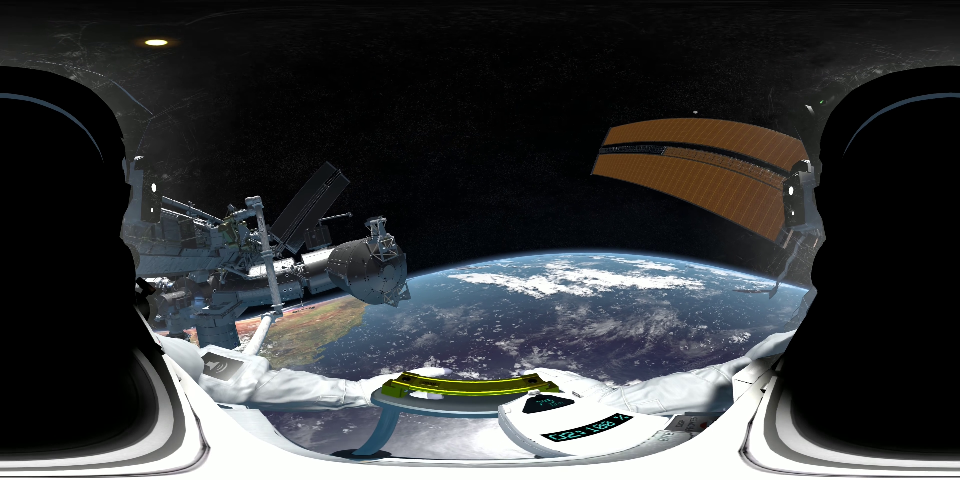}}
    		\centerline{\text{SpaceWalk 3min26s}}
    	\end{minipage}
    \begin{minipage}[h]{0.32\linewidth}
    		\vspace{3pt}
    		\centerline{\includegraphics[width=\textwidth]{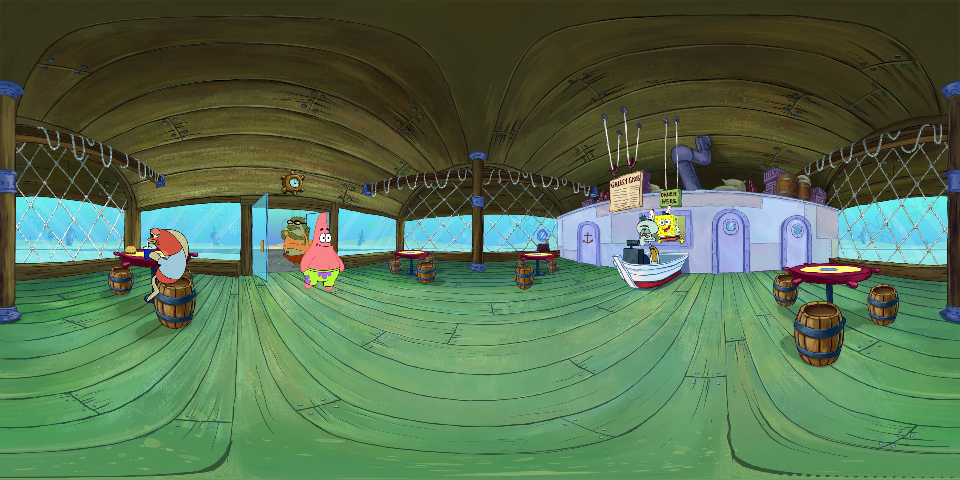}}
    	 
    		\centerline{\text{SpongeBob 53s}}
    	\end{minipage}
    \begin{minipage}[h]{0.32\linewidth}
    		\vspace{3pt}
    		\centerline{\includegraphics[width=\textwidth]{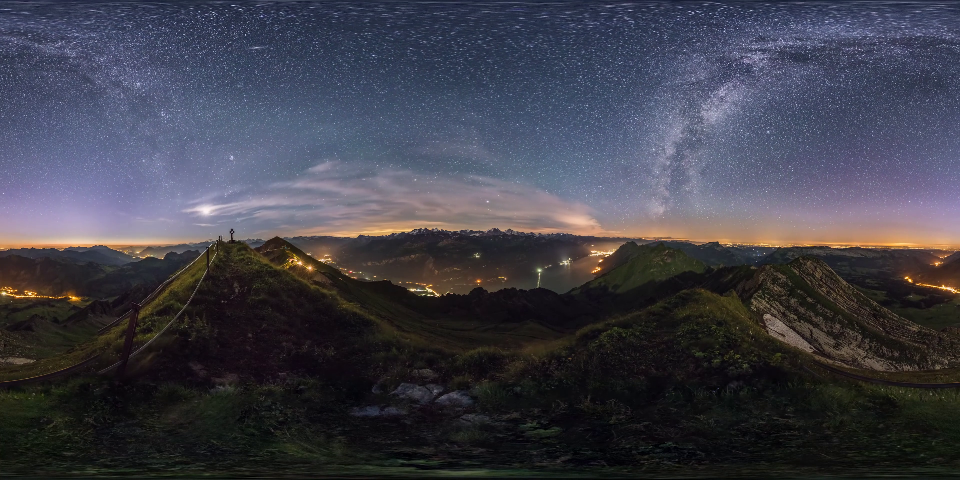}}
    	 
    		\centerline{\text{StarNight 26s}}
    	\end{minipage}
    \begin{minipage}[h]{0.32\linewidth}
    		\vspace{5pt}
    		\centerline{\includegraphics[width=\textwidth]{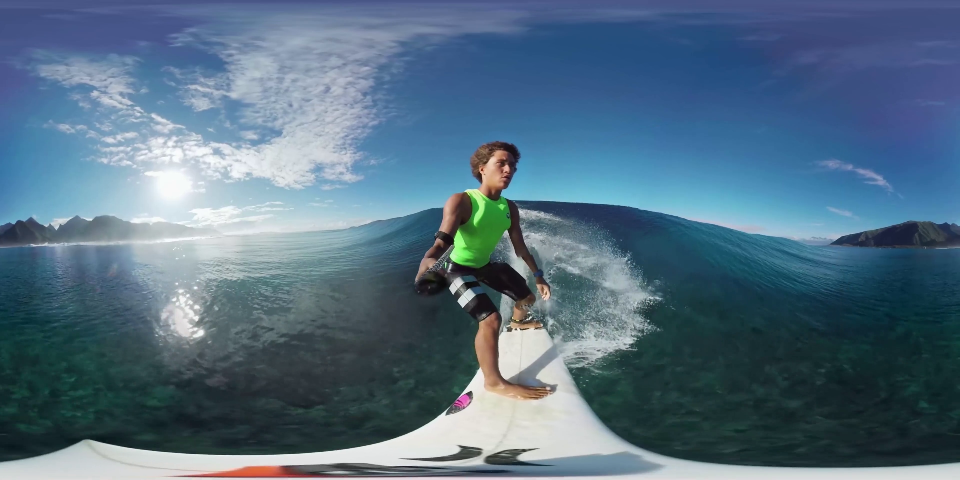}}
    	 
    		\centerline{\text{Surfing 1min4s}}
    	\end{minipage}
    \begin{minipage}[h]{0.32\linewidth}
    		\vspace{3pt}
    		\centerline{\includegraphics[width=\textwidth]{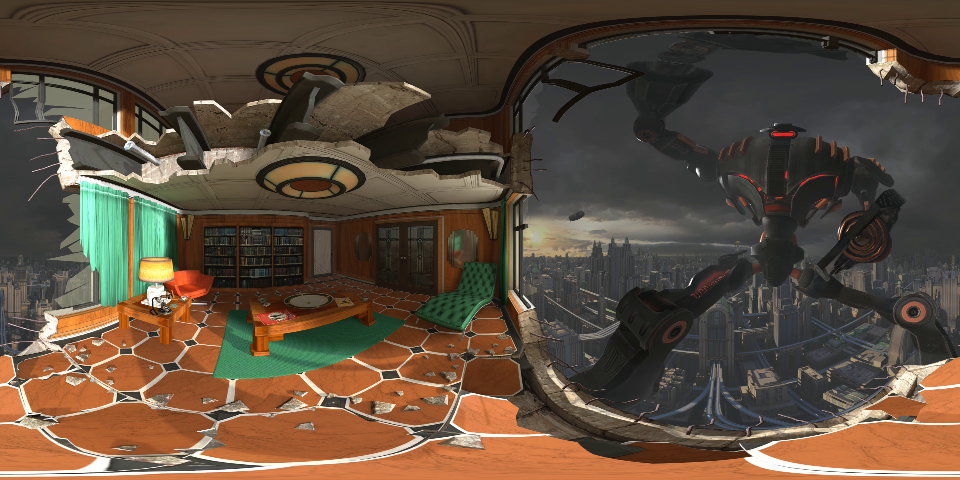}}
    	 
    		\centerline{\text{UFO 26s}}
    	\end{minipage}
    \begin{minipage}[h]{0.32\linewidth}
    		\vspace{3pt}
    		\centerline{\includegraphics[width=\textwidth]{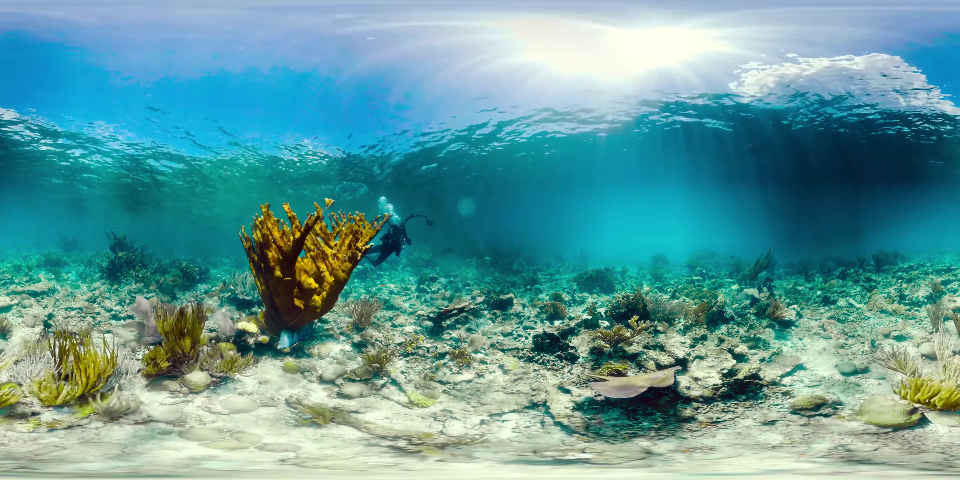}}
    	 
    		\centerline{\text{UnderWater 4min36s}}
    	\end{minipage}
	\caption{Sample images from each video}

\end{figure}
\subsection{Participant}
Table \ref{tab:participants} presents the demographic profile of all the participants. Our experiment has a total of 50 participants, with an even gender distribution of 25 males and 25 females. Half of the participants are between 20 and 25 years old and the remaining participants are sampled from other representative age brackets. 20 participants have experienced of VR devices before and three of them use VR frequently. Notably, 43 out of the 50 participants, representing approximately 86\% of the total, have experience (include heard, sometimes and frequently use) of VR. Only seven participants, the majority of whom are the elderly, have never experienced VR.

\begin{table}[htbp]
\centering
\caption{Participants Information}
  \label{tab:participants}

\hspace*{-6.0cm} 
\begin{minipage}[t]{0.45\textwidth}
\centering
\begin{tabular}{cc}
    \toprule
    \multicolumn{2}{c}{\textbf{Gender}} \\  
    \midrule
    Male & Female \\ 
    \midrule
    25 & 25 \\
    \bottomrule
\end{tabular}
\vfill 
\hspace*{3cm} 
\vspace*{-7cm} 

\end{minipage}\hfill

\begin{minipage}[t]{0.48\textwidth}
\hspace*{1cm} 
\vspace*{-1cm} 
\centering
\vspace*{-0.465cm} 
\hspace*{2.5cm} 
\begin{tabular}{ccccc}
    \toprule
    \multicolumn{5}{c}{\textbf{Age}} \\  
    \midrule
   \textless{}20 & $20 - 25$ &  $26 - 35$ & $36 - 45$ &  \textgreater{}45 \\ 
    \midrule
    5 & 25 & 6 & 7 & 7 \\
    \bottomrule
\end{tabular}
\end{minipage}

\vspace*{0.5cm} 

\begin{minipage}{\textwidth}
\hspace*{-9.2cm} 
\centering
\begin{tabularx}{0.49\textwidth}{>{\centering\arraybackslash}X>{\centering\arraybackslash}X>{\centering\arraybackslash}X>{\centering\arraybackslash}X}
    \toprule
    \multicolumn{4}{c}{\textbf{VR Experience}} \\  
    \midrule
     Never & Heard & Sometimes & Frequently \\ 
    \midrule
    7 & 20 & 20 & 3  \\
    \bottomrule
\end{tabularx}
\end{minipage}

\end{table}
\section{USER BEHAVIOR ANALYSIS}

Due to the lack of recognized ground truth for saliency maps, the complexity of gaze data and the inevitable abnormal data caused by blinking, head fixation is still often used as a proxy for gaze attention. This approach can be considered as an approximation, assuming that the head motion is a sufficient proxy for gaze attention. Researchers often use Gaussian filter \cite{atsal,dhp} to convolve each fixation point to generate saliency maps. However, a report issued by the Hong Kong government \cite{HK} and research by Rai et al. \cite{7965659}  both confirm that there is an offset angle between the center of the FoV $H$ and gaze fixation $G$ .

So, how much confidence do we have in the hypothesis that the center of the FoV $H$ is a reasonable reflection of gaze fixation $G$? In order to analyze user behavior patterns, we first perform an analysis of the offset between the center of the FoV $H_{t}$ (the subscript \textit{t} represents time) and the gaze fixation point  $G_{t}$ (for convenience, hereafter referred to as the relative gaze offset). Additionally, due to the equirectangular projection distortion, the $H_{t}$ and $G_{t}$ would have to be mapped to the latitude coordinate (the higher the latitude, the longer the distance), which will introduce errors to the relative gaze offset when calculating the Euclidean distance directly. So, we chose to calculate the angular difference between $H_{t}$ and $G_{t}$. We denote $Angle_{\textit{offset}}$ to represent the relative gaze offset, which is calculated by $Angle_{\textit{offset}} =  < H_{t} , G_{t} > $.

\begin{figure}[htbp]
    \centering
    \includegraphics[width=0.93\hsize]{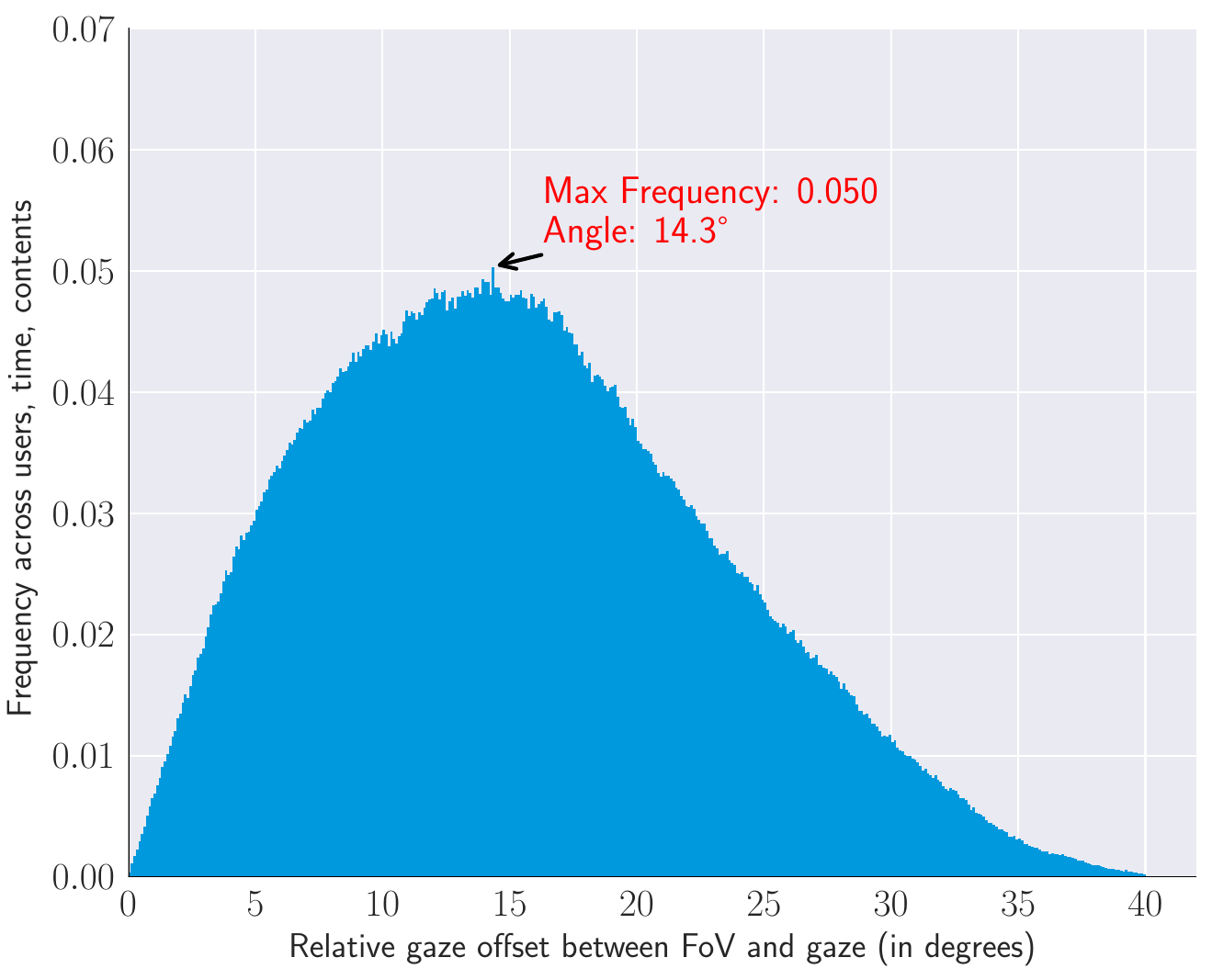}
    \caption{Frequency of the relative gaze offset (in degrees), across all users, time and contents }
    \label{fig:frequency}
\end{figure}

\begin{figure}[htbp]
    \centering
    \includegraphics[width=0.93\hsize]{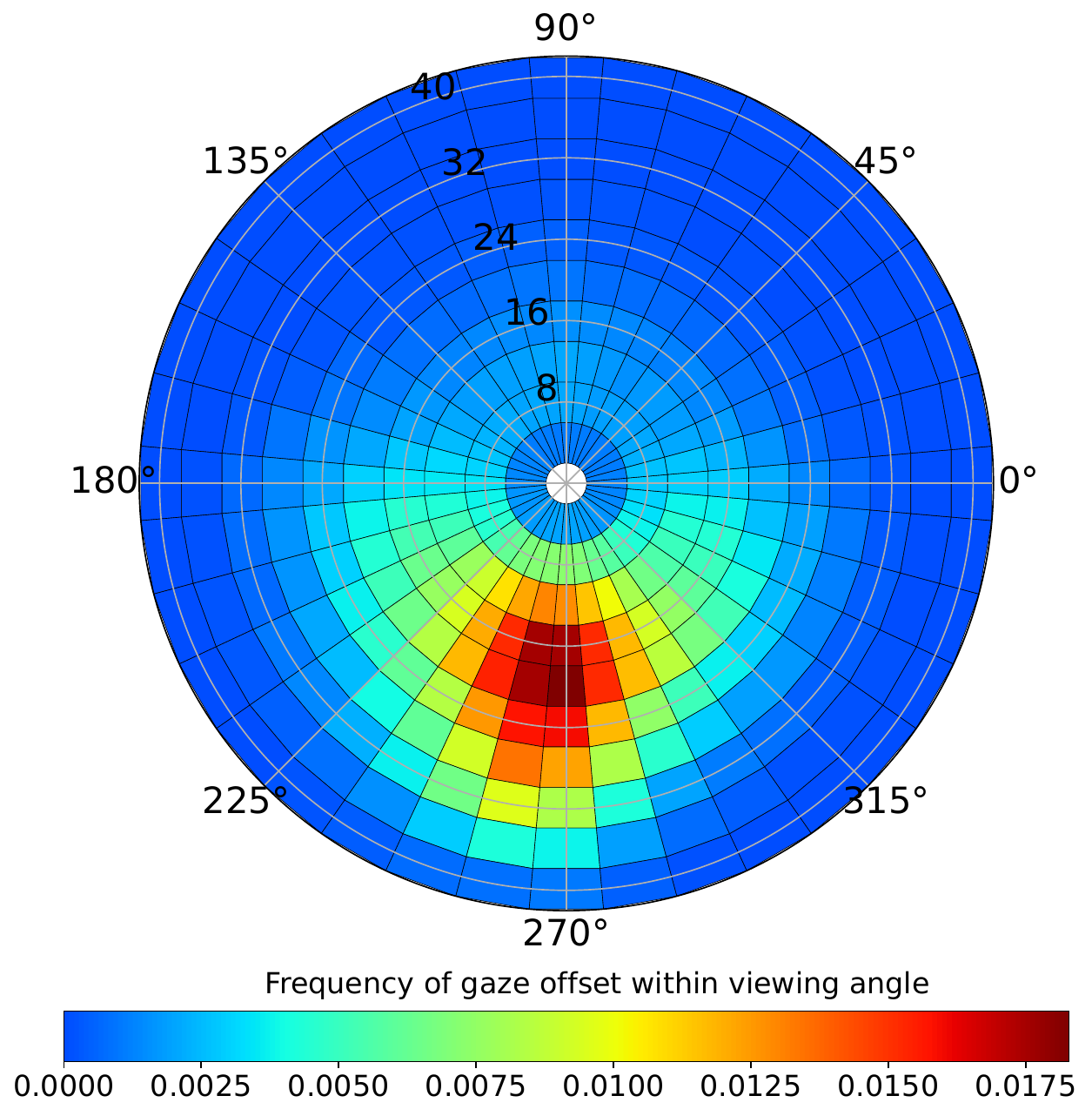}
    \caption{Gaze offset plotted as viewing direction ( polar angle bin is 10 degrees) and radial magnitude ( radial bin is 4 degrees).}
    \label{fig:polar}
\end{figure}

We track the head and gaze movements of 50 users in the dataset and calculate the relative gaze offset. The frequency distribution of the relative gaze offset $Angle_{\textit{offset}}$ is shown in Figure \ref{fig:frequency}, and we obtain results similar to the study \cite{7965659}. As shown in the Figure \ref{fig:frequency}, the curve is positively skewed and the relative gaze offset curve peaks at \textbf{14.3} degrees of visual angle from the center. Contrary to the common assumption that users always look at the center of FoV, their gaze often deviates from the direction of the head. Consequently, \textbf{we should be cautious when applying Gaussian weighting centered on the FoV to simulate the gaze fixation.} Accompanied by the question of whether this offset exists in all directions, we continue to try verifying another conclusion of Rai's: the gaze offset $Angle_{\textit{offset}}$ exhibits isotropic tendency within the panoramic scenario. We also plot the frequency of $Angle_{\textit{offset}}$ in polar coordination with the polar angle calculated by $\arctan\left(\frac{E_{y}}{E_{x}}\right)$ in Figure \ref{fig:polar} which represents the gaze direction relative to head. We are surprised to find that, unlike the result of previous studies based on panoramic images, users not only explore in all directions in panoramic scenario but also prefer to look at the lower position relative to the head direction, with a downward angle which is close to the peak of the relative gaze offset (i.e., 14.3 degree). It is not necessarily due to people's preference to look lower, but rather the result of the inherent structure of the human eye \cite{HK}. In summary, Figure \ref{fig:frequency} and Figure \ref{fig:polar} illustrate the deviation phenomenon of gaze fixation in the viewport from the perspective of probability and spatial distribution, respectively. 

We observe this behavior above all users and rule out any system error that could have been caused by calibration. We believe that this phenomenon may be due to a combination of factors, including users' gaze attention to the content of panoramic videos, where significant content is more likely located at lower positions in the FoV and the inherent structure of the human eye. Based on the above analysis, we propose this saliency weighting called \textit{Panonut}, which looks like a "donut" shape overall, and at the same time, relatively higher in the downward direction. That's why we named our dataset \textit{Panonut360}. 

\section{SALIENCY VISUALIZATION}

In this section, we provide pre-generated saliency maps and generation scripts in the dataset, following the work of Nguyen \textit{et al.} \cite{nguyen2019saliency}. Saliency maps were generated per second of the video by convolving each fixation point (for all observers of one video) with a Gaussian filter ($\sigma = 15^{\circ}$) \cite{15ms} around the gaze fixation for generating saliency visualization. Note that this is the process of generating ground truth for video saliency maps where users are actually focusing, not just an approximation with head tracking data. We chose several typical videos saliency ground truth to visualize in Figure 6.  Dynamic videos such as \textit{CutRope} have clear points of interest that capture users' attention, while static videos such as \textit{Cooking} also have a few fixed locations where users tend to focus their attention. For such videos, allocating bitrate based on saliency detection is a good choice. For first-person videos such as \textit{RollerCoaster}, users face forward to enhance their immersion experience, except for the initial scene exploration. For landscape videos such as \textit{StarNight}, users do not focus their attention on a particular area, instead, they explore the scenario freely. The above results indicate that the diversity of video content can have a significant impact on user behaviour. Our dataset can play a role in understanding the guidance of video content and audio to users.

\begin{figure}[htbp]
\label{visual}
\begin{minipage}{0.32\linewidth}
		\vspace{3pt}
		\centerline{\includegraphics[width=\textwidth]{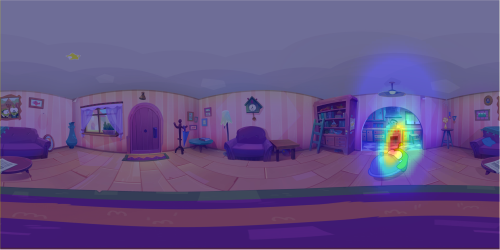}}
		\centerline{\text{CutRope 16s}}
          
	\end{minipage}
\begin{minipage}{0.32\linewidth}
		\vspace{3pt}
		\centerline{\includegraphics[width=\textwidth]{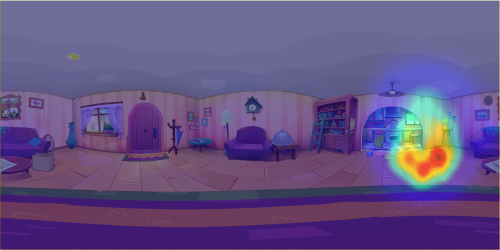}}
		\centerline{\text{CutRope 1min6s}}
           
	\end{minipage}
\begin{minipage}{0.32\linewidth}
		\vspace{3pt}
		\centerline{\includegraphics[width=\textwidth]{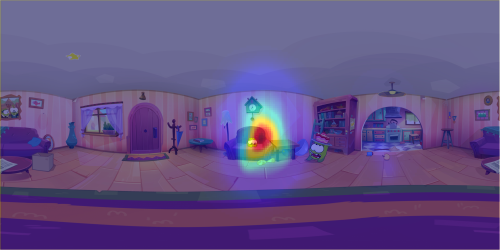}}
		\centerline{\text{CutRope 1min27s}}
            
	\end{minipage}
 \\
\begin{minipage}{0.32\linewidth}
		\vspace{3pt}
		\centerline{\includegraphics[width=\textwidth]{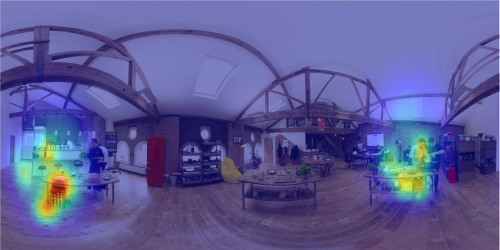}}
		\centerline{Cooking 49s}
           
	\end{minipage}
\begin{minipage}{0.32\linewidth}
		\vspace{3pt}
		\centerline{\includegraphics[width=\textwidth]{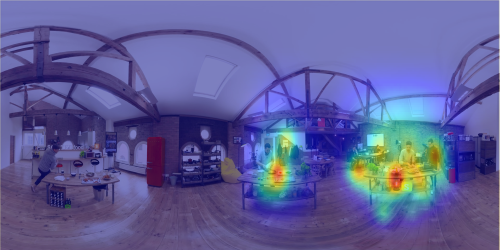}}
		\centerline{Cooking 1min27s}
            
	\end{minipage}
\begin{minipage}{0.32\linewidth}
		\vspace{3pt}
		\centerline{\includegraphics[width=\textwidth]{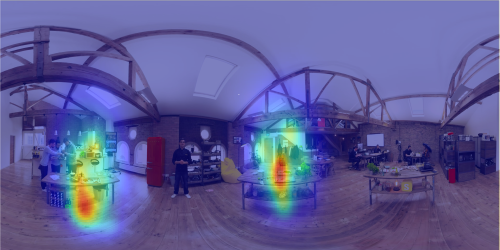}}
		\centerline{Cooking 3min7s}
            
	\end{minipage}
 \\
 \begin{minipage}{0.32\linewidth}
		\vspace{3pt}
		\centerline{\includegraphics[width=\textwidth]{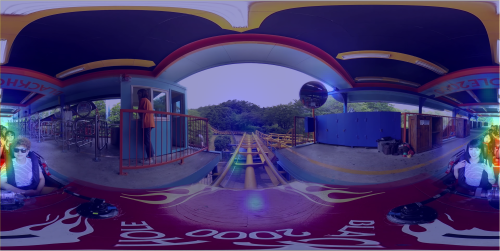}}
		\centerline{\text{RollerCoaster 6s}}
           
	\end{minipage}
\begin{minipage}{0.32\linewidth}
		\vspace{3pt}
		\centerline{\includegraphics[width=\textwidth]{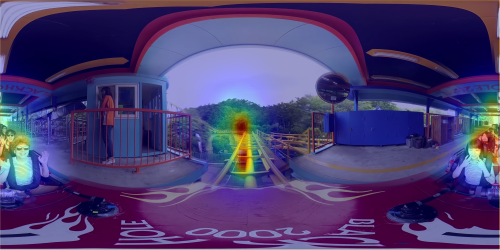}}
		\centerline{\text{RollerCoaster 8s}}
          
	\end{minipage}
\begin{minipage}{0.32\linewidth}
		\vspace{3pt}
		\centerline{\includegraphics[width=\textwidth]{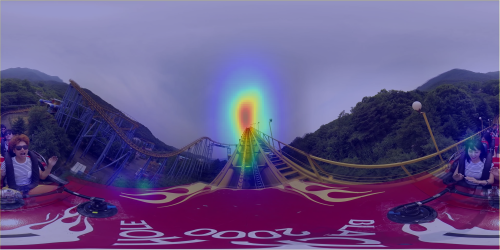}}
		\centerline{\text{RollerCoaster 50s}}
            
	\end{minipage}
 \\
 \begin{minipage}{0.32\linewidth}
		\vspace{3pt}
		\centerline{\includegraphics[width=\textwidth]{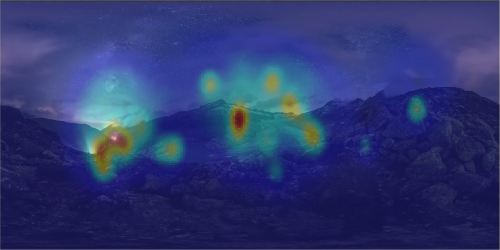}}
		\centerline{\text{StarNight 59s}}
           
	\end{minipage}
\begin{minipage}{0.32\linewidth}
		\vspace{3pt}
		\centerline{\includegraphics[width=\textwidth]{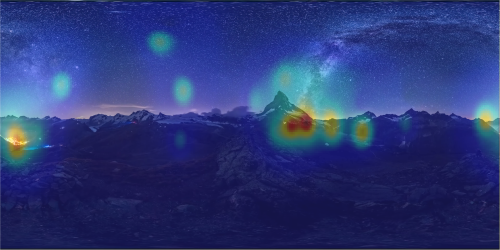}}
		\centerline{\text{StarNight 1min20s}}
            
	\end{minipage}
\begin{minipage}{0.32\linewidth}
		\vspace{3pt}
		\centerline{\includegraphics[width=\textwidth]{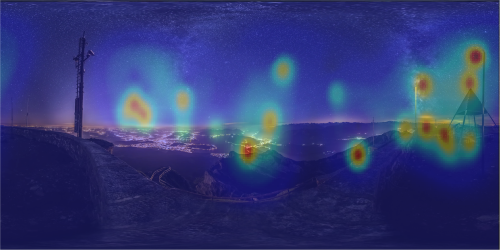}}
		\centerline{\text{StarNight 1min59s}}
            
	\end{minipage}
 
	\caption{Saliency visualization}
	
\end{figure}

\section{DATASET APPLICATION AND RELATED WORK}
Our dataset is versatile and can be used in numerous applications. In this section, we outline some potential use cases in tile-based panoramic video adaptation streaming and discuss related studies.


\subsection{FoV Prediction}


The FoV prediction tasks, which are typically used to make bitrate allocation decisions, have certain limitations\cite{graf2017towards}. Conventional representative methods for FoV prediction include filtering algorithms and RNN-based models (e.g., LSTM)\cite{Flare,EPASS360}. While RNN-based models often suffer from gradient disaster in long-term prediction. According to the best of our knowledge, Transformer has shown superior performance in capturing long-term dependence than RNN models, which means it is highly suitable for handling panoramic video downstream AI tasks that require temporal consistency, such as time series forecast. Existing studies have applied Transformer-based models to such tasks, e.g., \textit{Informer} \cite{zhou2021informer}. 


\subsection{Saliency Detection}
An accurate saliency detection model has played a key role in tile-based panoramic video adaptation and many other applications, such as ROI adaptive video coding \cite{grois2010roi} and video object segmentation \cite{VideoSeg}. The general approach to saliency detection is to build supervised deep learning models based on backbone networks such as Convolutional Neural Network (CNN)\cite{Unique}. In summary, since the saliency detection model has a significant impact on QoE, expanding the saliency dataset is beneficial.
\section{CONCLUSION}

In this paper we introduce the \textit{Panonut360} dataset of 50 users recorded while they were watching panoramic videos using the HTC VIVE Pro HMD. The dataset includes over 60 hours of data collected at 120Hz which combines multidimensional information. We then present examples of statistics that can be extracted from the dataset to analyze user viewing behaviour and pre-generated saliency maps. We firmly believe that this dataset will be comprehensive resource for researchers, empowering them to develop better adaptive streaming systems for panoramic videos. 

\section{ACKNOWLEDGMENTS}

This work was supported in part by the National Natural Science Foundation of China (NSFC) with Grant 62271224 and in part by the ZTE Industry-University-Institute Cooperation Funds under Grant No. IA20230728008.

\printbibliography
\end{document}